# Towards solving the multiple extension problem: combining defaults and probabilities

Eric Neufeld and David Poole
Logic Programming and Artificial Intelligence Group
Department of Computer Science
University of Waterloo
Waterloo, Ontario, Canada

**Abstract**

The *multiple extension* problem frequently arises in both diagnostic and default reasoning. That is, in many settings it is possible to use any of a number of sets of instances defaults or hypotheses to explain (expected) observations. In some cases, we choose among explanations by making inferences about information believed to be implicit in the problem statement. If this isn't possible, we may still prefer one explanation to another because it is more likely or optimizes some other measureable property: cost, severity, fairness.

We combine probabilities and defaults in a simple unified framework that retains the logical semantics of defaults and diagnosis as construction of explanations from a fixed set of possible hypotheses. We view probability as a *property* of an explanation that can be computed from a what is known and what is hypothesized by a valuation function. We present a procedure that performs an iterative deepening branch-and-bound search for explanations with the property that the first path found is the most likely. The procedure does not consider unlikely paths until more likely ones have been eliminated.

We outline a way in which probabilities are not constrained by *a priori* independence assumptions; rather, these statistical assumptions are set up as defaults.

While we use probability as a way of preferring one answer to another, the results apply to any property of an explanation having a valuation function meeting some usefulness criteria.

## 1 Introduction

In many situations, we want to use generalized knowledge, prototypical knowledge or knowledge not always true in our domain.

Examples include:

- default reasoning, where we use an instance of a default in order to predict some proposition, if we can do so consistently [27,22].

- diagnosis, where we hypothesize problems with a system that explain observed malfunctions and are consistent with what we know about the system [26,28,6].

- natural language understanding, where given some discourse we want to compute the underlying presuppositions to allow for the most meaningful reply [17,20].

- planning, where we hypothesize a sequence of actions that imply completion of a goal[12].

- user modelling, where we hypothesize about the knowledge and beliefs of a user[30].

Here, we consider a use of logic different from just finding what deductively follows from our knowledge. We use the idea of constructing explanations from a fixed set of possible hypotheses supplied by the user. We try to build an explanation of our goal from instances of these possible hypotheses that is consistent with all our knowledge. If the possible hypotheses are defaults and the goal is something we are trying to predict, then this characterizes a natural form of default reasoning [24]. If the possible hypotheses are possible diseases or malfunctions of some system and the goal is the observed behaviour of the system then we have model-based diagnosis [11]. In [19] this kind of diagnostic reasoning is called *abduction* or *parsimonious covering theory* with the restriction that there is inferential distance one between observations and hypotheses. Similarly, recognition can be seen as constructing explanations for incomplete impressions where the possible hypotheses are the prototypes of the things we are trying to recognize.

In essence we are proposing something like scientific theory formation, but without the problem of generating the hypotheses. We assume that the user provides the forms of the hypotheses she is prepared to accept as part of an explanation.

Unless observations together with facts contain complete information about the system, it is natural that a single set of observations may have many explanations: hence, we get the "multiple extension" problem [8,14]. To choose among explanations, or extensions, we must define appropriate *comparators*. Existing comparators seem to fall into two broad classes. The first class uses information believed to be implicit in the problem statement to choose among explanations for the one that is correct in the intended interpretation. For example:

- Poole [23] proposes the notion of *most specific theory* to give semantics to the idea that when we have generalized knowledge and more specific knowledge represented as defaults, then we prefer to use the more specific knowledge when both are applicable[29]. To illustrate, suppose we believe the generalizations that mammals don't fly, bats fly, but dead bats do not. These can be represented as possible hypotheses. If Dracula is a dead bat, we want to predict that he doesn't fly with the explanation that dead bats don't fly rather than the explanation that mammals don't fly or predict that he flies because bats typically do fly; we prefer the explanation that uses the most specific information.

- Applied to the Yale shooting scenario[14,12], constructing explanations produces undesirable yet consistent theories that rankle intuition. (Other formalisms fare no better.) Goodwin[12] and Kautz[15] use *persistence* to choose the explanation that they believe is correct in the intended interpretation.

These approaches try to provide general ways to *complete* databases of predicates by making assumptions about the knowledge.

The second class of comparators views all consistent explanations as possibilities and so compares them on the basis of some measurable property. We call comparators of this sort *heuristic* since they require other *kinds* of knowledge about the domain. In our current research, we characterize this extra knowledge with a valuation function which assigns "goodness" values to competing explanations. Presently we are most interested in using likelihood to compare explanations given some observations, but in other situations, we might use cost or virulence.

This seems like an appropriate way to use probability in the commonsense domain. It has philosophical counterparts [3,16] and is similar in spirit to the implementations of [6] and [19].

A current investigation of this second class of comparators has yielded promising initial results. We give semantics for explainability and add semantics for comparators that are useful in setting where observations are known to be true, for example, diagnosis. We discuss properties a valuator must have so that a comparator is *useful*. In particular, probability is a useful comparator. We then describe a procedure that generates theories according to the order defined by the comparator. Lastly, we touch on a way to give semantics to valuators, (in particular, probability) with the aim of achieving efficiency but not at the cost of losing precision of meaning.

## 2 Semantics

### 2.1 Constructing explanations

Here we give the formal semantics of explainability.

The user provides the system with two sets of formulae:

$F$ is a set of closed formulae called *facts*.

$\Delta$ is a set of formulae called *possible hypotheses*. These can be defaults, prototypes, possible malfunctions or anything else we are prepared to accept as part of an explanation of some goal.

A goal $g$ is *explainable* if there is some $D$, a set of ground instances of members of $\Delta$, such that

$$F \cup D \models g$$
$F \cup D$ is consistent.

and $D$ is called an *explanation* of $g$. We sometimes say that $D$ *predicts* $g$ or that $D$ is the *theory that explains $g$*.

If $\Delta$ contains generalized knowledge ("birds fly", for example) and $g$ is some proposition we want to predict, we have default reasoning. If $\Delta$ contains possible diseases or malfunctions as well as defaults, and $g$ is the observations of the system, we have diagnosis.

Poole describes this system in the framework of the first order predicate logic, but notes that the system is not restricted to that logic[21]. We want an explanation which predicts the goal, but does not predict anything which we know is false.

## 2.2 Comparing explanations

A *valuator* is a function $m(g, D)$ where $D$ is an explanation of goal $g$. We assume the set of facts $F$ fixed in any application. The range of $m$ is a poset ordered by $\preceq$. We *prefer* explanation $D_1$ to explanation $D_2$ if $m(g, D_2) \preceq m(g, D_1)$ and we write $D_2 \preceq D_1$. If $D_1 \preceq D_2$ and $D_2 \preceq D_1$ then $D_1 \approx D_2$. If $D_1 \preceq D_2$ and not $D_1 \approx D_2$ then $D_1 \prec D_2$.

A valuator $m$ is *useful* if:

1. $m$ is defined at least for every $D$ and $g$ such that D is an explanation for $g$. This ensures that $m$ is defined whenever we have two explanations to compare, but doesn't insist that $m$ be defined when $D$ is irrelevant to $g$. This to some extent accommodates the proponents of intuitive probability who do not insist that $h|e$ be defined for all $h$ and $e$.

2. if $F \cup D_2 \Rightarrow D_1$, then $m(g, D_2) \preceq m(g, D_1)$. This corresponds to preferring explanations which make fewer assumptions, in the sense of logical implication or the subset relation (the above implication holds, in particular when $D_1 \subseteq D_2$). Following William of Occam, we always prefer a simpler explanation. Adding hypotheses decreases certainty or increases cost, both undesirable features.

These assumptions are consistent with Aleliunas' extension[1] of the Cox's result [5]. Then it is easy to see that standard real-valued probability is a useful valuation function and $m$ could be interpreted as the conditional probability of $D$ given $g$. We retain the more general notion of valuation of explanations - it seems consistent with every heuristic comparator we can think of.

Below, we assume that $m$ is a useful valuator. If $D_k$ is such that there is no $j$ such that $m(D_k, g) \prec m(D_j, g)$, then $D_k$ is a *preferred explanation* with respect to $m$ and $g$ and $F$. We usually simply call $D_k$ a preferred explanation when the context is clear.

## 3 Implementation

Some diagnostic programs [7] compute the probability of all possible elements of $\Delta$ given the observations and let the user select the best. This is only possible in specialized domains. We can also find a preferred explanation by generating all explanations (using the machinery described below), then valuating and sorting them. We propose the following as an intelligent model of reasoning: to explain some observations, an agent will follow a particular line of thought until it becomes weakened by too many strong assumptions; then she will switch to a more plausible line of thought. Such an agent finds a best explanation without having to completely explore obscure possibilities in detail.

Poole, Goebel and Aleliunas describe an implementation of explainability in [25] and include a terse Prolog implementation. Briefly, the theorem prover tries to prove some observations $g$ from $F$ and $\Delta$, and makes $D$ the set of instances of members of $\Delta$ used in the proof, as long as it fails to prove that $F \cup D$ is inconsistent. When the theorem prover uses a ground instance $\delta$ of $\Delta$, we say it makes the assumption $\delta$ or assumes the hypothesis $\delta$. The theorem prover may also obtain new observations if there are predicates proved by querying the user.

We use the same technique, but build all proofs incrementally using a separate process for each proof. If $D$ only explains some of the observations in $g$, we compute $m$ by considering only the observations in $g$ explained by $D$.

Each partial proof of the observations, carries a state parameter $< O, D, N >$ describing currently known global observations $O$, a local partial explanation $D$ and the value $N = m(O, D)$. If we are ranking explanations by likelihood, $N$ is an optimistic upper bound on the final value of $m$ when the proof halts, i.e., if the unexplained observations follow from the facts and the current assumptions. The procedure begins with the initial observations, the null hypothesis and an appropriate initial value for the null hypothesis. In the case of real-valued probabilities, this is 1.0. $N$ decreases in any process that assumes a new hypothesis; it may change in any way in any proof whenever new observations are made.

The procedure follows:

1. Create a process with initial state $< g, \{\}, n >$ where $g$ represents the initial observations, the empty set represents the null explanation, and $n$ is an appropriate initial value.

2. If a process assumes $h$, create a child process that is a copy of the parent. If $F \cup D \not\models \neg h$, set the state of the parent process from $< O, D, N >$ to $< O, \{h\} \cup D, m(O, \{h\} \cup D) >$ and suspend it; else kill the process. Make the child backtrack to the last choice point (i.e., "un-assume" $h$).

3. If a new observation $o$ is made (if the user provides more information about the domain), reset the states of all processes from $< O, D, N >$ to $< \{o\} \cup O, D, m(\{o\} \cup O, D >$. Suspend all processes.

4. If all processes are suspended, kill all processes with inconsistent $O \cup D$ and restart a process with a highest value of $N$. If no running or suspended processes remain, there is no solution, halt. Otherwise, continue running the current process until either a new fact is learned or a new hypothesis is assumed, then go to step 2. If the current process completes a proof, print the answer and halt.

This is completely undecidable, since the procedure defines a control structure for a theorem prover that does consistency checks. However, if it halts, it has either found a preferred explanation or there is no explanation.

As long as no new observations are made, step 2 ensures that we are always working on the most promising explanation. Step 3 ensures all proofs exploit any new knowledge.

We have a prototype Prolog program which compiles specifications of $F$ and $\Delta$ and $m$ to a Prolog program that searches for a preferred explanation. With an empty $\Delta$ and a null valuator, the compiled code behaves exactly like Prolog.

## 4 Measuring properties of explanations

We have given semantics for explainability and preferability given a valuation function, but we have not given semantics for the valuation function itself. We are investigating giving ways of semantics to valuators based on the explainability formalism and give the basic ideas below with specific reference to probabilities.

The practical activity of calculating likelihood always seems to involve making statistical assumptions *in the absence of other information*. Otherwise the required calculations are intractable. For example:

- In the electronic circuit diagnosis problem, gates fail independently [6].

- Conditional independence of symptoms given diseases [2,19].

- The maximum entropy assumption. [4]

- When performing diagnosis, given no information to the contrary, we assume an individual might have contracted some disease with likelihood equal to the statistical mean unless we have information that the individual is particularly susceptible or particularly resilient to that disease.

In [13], Benjamin Grosof makes similar observations and places calculation of probabilities in the setting of Reiter's defaults and pointwise circumscription. He also discusses another independence assumption, maximization of conditional independence.

But it may not be natural to make any single statistical assumption an axiom of a theory of probability. A frequent problem with the old medical diagnostic systems, for example, is that a host of statistical assumptions were forced on the data *a priori*, resulting in the loss of useful exceptional information [10,18]. No statistical assumption is foolproof in general and there may be situations where we want to make other *kinds* of assumptions.

The explainability formalism potentially gives us a simple solution to all these problems. We divide a theory of probability into facts $F_p$ and *domain dependent* statistical assumptions $\Delta_p$. The facts contain user-supplied statistics as well as the standard axioms of probability theory. $\Delta_p$ may contain independence and other kinds of assumptions. The user or the theorem prover may decide what statistical assumptions are appropriate in a given setting. In any case, the assumptions are used only when no other information is known or can be derived. This also gives an interesting symmetry: probability as a useful valuator of explanations, can be used to guide the explainability machinery to produce a preferred explanation first and the machinery for explainability can be used to obtain the most meaningful probability in given situation.

Many uncertainty formalisms trade tractability for precision of meaning. We believe that it is the special case where the expensive computation occurs in our system, and in general we can have the best of both worlds. In any case, this idea seems to accommodate any valuation functions meeting our usefulness criteria and allows arbitrary exceptional information. Practical ways to flexibly incorporate such statistical assumptions are currently under investigation.

## 5 Conclusions

We have borrowed from the ideas of Reiter[28], deKleer[6], Reggia[26], Genesereth[11,9], and Grosof[13] to construct a framework for constructing explanations independent of the logic used (though we favour first order predicate logic) and independent of the heuristic used to guide the theorem proving procedure to the preferred explanation (though we favour probability and its derivatives).

We suggested that the probabilities could be calculated in a smaller version of the framework where statistical assumptions were modelled as possible hypotheses, providing an interesting duality to the system.

Our implementation handles diagnosis of simple circuits (the full adder example from Genesereth[11]), and small medical diagnostic problems. The initial results seem to suggest that probability and logic work together well as two simple different interacting cooperative processes; however, many interesting theoretical and practical problems remain to be solved.

## 6 Acknowledgements

We thank members of the LPAIG for discussions and suggestions, especially Romas Aleliunas, Randy Goebel, Denis Gagne and Scott Goodwin for constructive criticisms.

This research was supported by the Natural Science and Engineering Research Council of Canada and the Institute for Computer Research at the University of Waterloo.

# References

[1] Romas Aleliunas. Mathematical models of reasoning based on abstract probability algebras. 1986. manuscript.

[2] M. Ben-Bassat, R.W. Carlson, V.K. Puri, M.D. Davenport, J.A. Schriver, M. Latif, R. Smith, L.D. Portigal, E.H. Lipnick, and M.H. Wel. Pattern-based interactive diagnosis of multiple disorders: the MEDAS system. *IEEE Transactions on Pattern Analysis and Machine Intelligence*, 2:148–160, 1980.

[3] Rudolph Carnap. *Logical Foundations of Probability*. University of Chicago Press, 1950.

[4] Peter Cheeseman. A method of computing generalized Bayesian probability values for expert systems. In *Proceedings IJCAI-83*, pages 183–202, 1983.

[5] Richard T. Cox. Probability, frequency and reasonable expectation. *American Journal of Physics*, 14(1):1–13, 1946.

[6] Johan de Kleer. Diagnosing multiple faults. *Artificial Intelligence*, 32(1):97–130, 1987.

[7] F. T. deDomba. Computer-assisted diagnosis of abdominal pain. In J. Rose and J.H. Mitchell, editors, *Advances in Medical Computing*, pages 10–19, Churchill-Livingstone, New York, 1979.

[8] David W. Etherington and Raymond Reiter. On inheritance hierarchies with exceptions. In *Proceedings AAAI-83*, pages 104–108, 1983.

[9] J.J. Finger and Michael R. Genesereth. *RESIDUE: A Deductive Approach to Design Synthesis*. Technical Report HPP-85-1, Stanford Heuristic Programming Project, January 1985.

[10] Dennis G. Fryback. Bayes' theorem and conditional nonindepen'dence of data in medical diagnosis. *Computers and Biomedical Research*, 11:423–434, 1978.

[11] Michael R. Genesereth. The use of design descriptions in automated diagnosis. *Artificial Intelligence*, 24:411–436, 1984.

[12] Randy G. Goebel and Scott D. Goodwin. Applying theory formation to the planning problem. In F.M. Brown, editor, *The Frame Problem in Artificial Intelligence: Proceedings of the 1987 Workshop*, Morgan Kaufmann, Los Altos, Cal., April 1987.

[13] Benjamin N. Grosof. Non-monotonicity in probabilistic reasoning. In *Proceedings Uncertainty in Artificial Intelligence Workshop*, pages 91–98, August 1986.

[14] S. Hanks and D. McDermott. Default reasoning, nonmonotonic logic and the frame problem. In *Proceedings AAAI-86*, pages 328–333, 1986.

[15] H. Kautz. The logic of persistence. In *Proceedings AAAI-86*, pages 401–405, 1986.

[16] J.M. Keynes. *A Treatise on Probability*. MacMillan, London, 1921.

[17] R.E. Mercer and R. Reiter. *The Representation of Presuppositions Using Defaults*. Technical Report 82-1, University of British Columbia, Dept. of Computer Science, 1982.

[18] Marija A. Norusis and John A. Jacquez. Diagnosis. I. Symptom nonindependence in mathematic models for diagnosis. *Computers and Biomedical research*, 8:156–172, 1975.

[19] Yun Peng and James Reggia. Plausibility of diagnostic hypotheses: the nature of simplicity. In *Proceedings AAAI-86*, pages 140–145, 1986.

[20] C. Raymond Perrault. An application of default logic to speech act theory. unpublished draft.

[21] David Poole. Variables in hypotheses. 1987. to appear, Proceedings IJCAI-87.

[22] David L. Poole. *Default Reasoning and Diagnosis as Theory Formation*. Technical Report CS-86-08, University of Waterloo Department of Computer Science, 1986.

[23] David L. Poole. On the comparison of theories: preferring the most specific explanation. In *Proceedings IJCAI-85*, pages 144–147, 1985.

[24] David L. Poole. The use of logic. 1987. to appear in Computational Intelligence Special Issue on Taking Issues in Logic edited by Hector Levesque.

[25] David L. Poole, R.G. Goebbel, and R. Aleliunas. Theorist: a logical reasoning system for defaults and diagnosis. In Nick Cercone and Gordon McCalla, editors, *The Knowledge Frontier: Essays in the Representation of Knowledge*, Springer-Verlag, New York, 1987.

[26] James A. Reggia, Dana S. Nau, and Pearl Y. Wang. Diagnostic expert systems based on a set covering model. *International Journal of Man-Machine Studies*, 19:437–460, 1983.

[27] Raymond Reiter. A logic for default reasoning. *Artificial Intelligence*, 13:81–132, 1980.

[28] Raymond Reiter. A theory of diagnosis from first principles. *Artificial Intelligence*, 32(1):57–96, 1987.

[29] D. S. Touretzky. Implicit orderings of defaults in inheritance systems. In *Proceedings AAAI-84*, pages 322–325, 1984.

[30] Paul van Arragon. User models in theorist. 1986. Ph.D. Thesis Proposal, Department of Computer Science, University of Waterloo.